\documentclass[letterpaper, 10 pt, conference]{ieeeconf}  

\IEEEoverridecommandlockouts                              

\usepackage{graphics} 
\usepackage{epsfig} 
\usepackage{mathptmx} 
\usepackage{times} 
\usepackage{amsmath} 
\usepackage{amssymb,bm}  
\usepackage{color}

\usepackage[style=ieee]{biblatex}

\DeclareSourcemap{
  \maps{
    \map{
      \pertype{article}
      \step[fieldset=url, null]
      \step[fieldset=doi, null]
      \step[fieldset=issn, null]
      \step[fieldset=isbn, null]
      \step[fieldset=note, null]
      \step[fieldset=editor, null]
      \step[fieldset=urldate, null]
      \step[fieldset=file, null]
    }
  }
}
\DeclareSourcemap{
  \maps{
    \map{
      \pertype{inproceedings}
      \step[fieldset=url, null]
      \step[fieldset=doi, null]
      \step[fieldset=issn, null]
      \step[fieldset=isbn, null]
      \step[fieldset=note, null]
      \step[fieldset=editor, null]
      \step[fieldset=urldate, null]
      \step[fieldset=file, null]
    }
  }
}
\DeclareSourcemap{
  \maps{
    \map{
      \pertype{incollection}
      \step[fieldset=url, null]
      \step[fieldset=doi, null]
      \step[fieldset=issn, null]
      \step[fieldset=isbn, null]
      \step[fieldset=note, null]
      \step[fieldset=editor, null]
      \step[fieldset=urldate, null]
      \step[fieldset=file, null]
    }
  }
}

\title{\LARGE \bf
Non-impulsive Contact-Implicit Motion Planning \\for Morpho-functional Loco-manipulation
}

\author{Adarsh Salagame$^{1\text{\textdagger}}$, Kruthika Gangaraju$^{1\text{\textdagger}}$, Harin Kumar Nallaguntla$^{1}$,\\ Eric Sihite$^{2}$, Gunar Schirner$^{1}$, Alireza Ramezani$^{1*}$
\thanks{$^{1}$This author is with the Department of Electrical and Computer Engineering, Northeastern University, Boston MA
        {\tt\small salagame.a, gangaraju.k, nallaguntla.h, G.Schirner, a.ramezani@northeastern.edu*}}%
\thanks{$^{2}$ This author is with California Institute of Technology, Pasadena CA
		{\tt\small esihite@caltech.edu}}%
\thanks{\text{\textdagger}These authors have equal contribution to this work.}
\thanks{$*$Indicates the corresponding author.}
}

\newcommand{\e}[1]{\times 10^{#1}}

\begin{document}

\maketitle
\thispagestyle{empty}
\pagestyle{empty}

\begin{abstract}
Object manipulation has been extensively studied in the context of fixed base and mobile manipulators. However, the overactuated locomotion modality employed by snake robots allows for a unique blend of object manipulation through locomotion, referred to as loco-manipulation. The following work presents an optimization approach to solving the loco-manipulation problem based on non-impulsive implicit contact path planning for our snake robot COBRA. We present the mathematical framework and show high fidelity simulation results for fixed-shape lateral rolling trajectories that demonstrate the object manipulation.
\end{abstract}

\section{INTRODUCTION}
\label{sec:intro}

Snake locomotion encompasses various techniques tailored for different environments and challenges. Lateral undulation, exemplified in works such as \cite{wiriyacharoensunthorn_analysis_2002, ma_analysis_2003, ma_analysis_1999}, relies on anisotropic friction to propel the snake forward in a sinusoidal pattern. Rectilinear motion, as described in \cite{rincon_ver-vite_2003, ohno_design_2001}, involves the compression and expansion of scales for longitudinal movement, ideal for navigating tight spaces. Sidewinding gait, demonstrated in studies like \cite{liljeback_modular_2005, burdick_sidewinding_1994}, is deployed on slippery or sandy terrains, employing a sinusoidal gait for lateral motion. In confined spaces, snakes utilize the concertina gait, as outlined in \cite{shan_design_1993}, involving coiling and uncoiling to progress longitudinally. Additionally, non-snakelike gaits such as the inchworm gait, slinky gait, lateral rolling gait, and tumbling locomotion, proposed in works like \cite{yim_new_1994, rincon_ver-vite_2003}, leverage the articulated nature of the snake's body for unique locomotion patterns.

Research in snake robotics has predominantly centered on emulating snake locomotion and replicating the distinctive movement patterns of biological snakes. However, the inherent redundancy and highly articulated nature of snake bodies have often been overlooked for manipulation. 

This study aims to explore the manipulation capabilities afforded by the redundant body structures of snake robots, particularly for interacting with objects in the environment. Addressing this contact-rich problem presents intriguing prospects for leveraging contact-implicit optimization, which represents a prevalent design paradigm in the field of locomotion and for unknown reasons is less explored in snake-type robots \cite{sihite_multi-modal_2023,ramezani_atrias_2012,sihite_efficient_2022,dangol_control_2021}.

\begin{figure}
    \centering
    \includegraphics[width=1.0\linewidth]{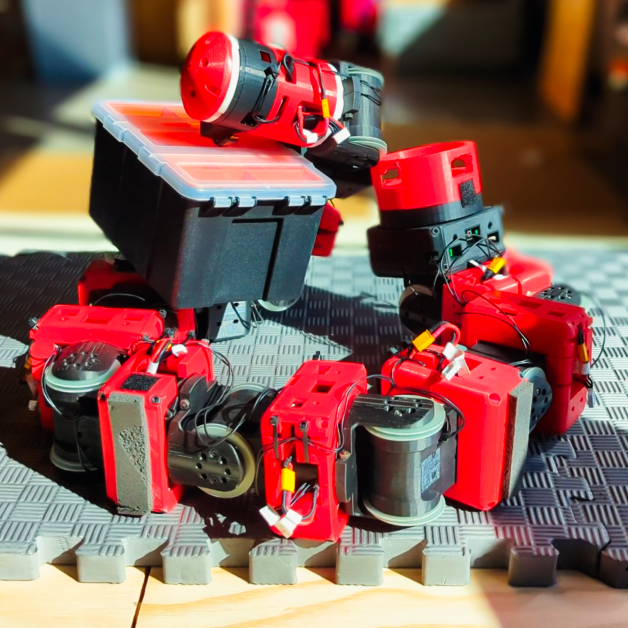}
    \caption{Illustrates loco-manipulation problem concerning carrying a box with COBRA}
    \label{fig:cover-image}
\end{figure}

\section{Quick Overview of COBRA, Objectives, Paper Contributions}
\label{sec:motiv}

As depicted in Fig.~\ref{fig:cover-image}, the COBRA system comprises eleven actuated joints. The frontal section of the robot houses a head module housing the onboard computing system, a communication module, and an inertial measurement unit (IMU) for navigation. At the rear, an interchangeable payload module accommodates additional electronics tailored to specific tasks undertaken by COBRA. The remaining components consist of identical modules, each containing a joint actuator and a battery.

A latching mechanism in the head module allows for the attachment of a gripper, facilitating object manipulation through a more conventional grip-based approach. This mechanism features a Dynamixel XC330 actuator situated within the head module, driving a central gear. This gear interfaces with partially geared sections of four fin-shaped latching fins. The curved outer face of each latching fin spans an arc length equal to 1/4 of the circumference of the head module's circular cross-section. When retracted, these four fins form a thin cylinder coinciding with the cylindrical face of the head module. A dome-shaped cap positioned at the end of the head module accommodates the fins between it and the main body of the head module, with clevis pins used to secure the fins in place.

The primary research objectives include:

\begin{itemize}
    \item Understanding efficient modeling techniques conducive to complex locomotion and object manipulation by COBRA.
    \item Exploring optimal control design approaches to effectively move the joints along desired trajectories for manipulating objects.
\end{itemize}

The key contributions of this work entail the proposal of an optimization approach based on non-impulsive implicit-contact path planning for COBRA. We demonstrate that this method can yield optimal joint trajectories for desired object movements over flat surfaces, showcasing its utility through simulation.
 
This paper is structured as follows: Initially, we introduce the fundamental concept underpinning the motion optimization approach employed in this study, detailing the incorporation of contact forces within the context of object locomotion and manipulation. Subsequently, we outline the simulation parameters utilized, followed by the presentation of simulated outcomes encompassing various joint motions, including object manipulation employing sidewinding and lateral rolling gaits of different shapes (C-, S-, J-shaped). Finally, we conclude with concise closing remarks.

\section{Motion Optimization}
\label{sec:mo-opt}

\begin{figure}
    \centering
    \includegraphics[width=0.9\linewidth]{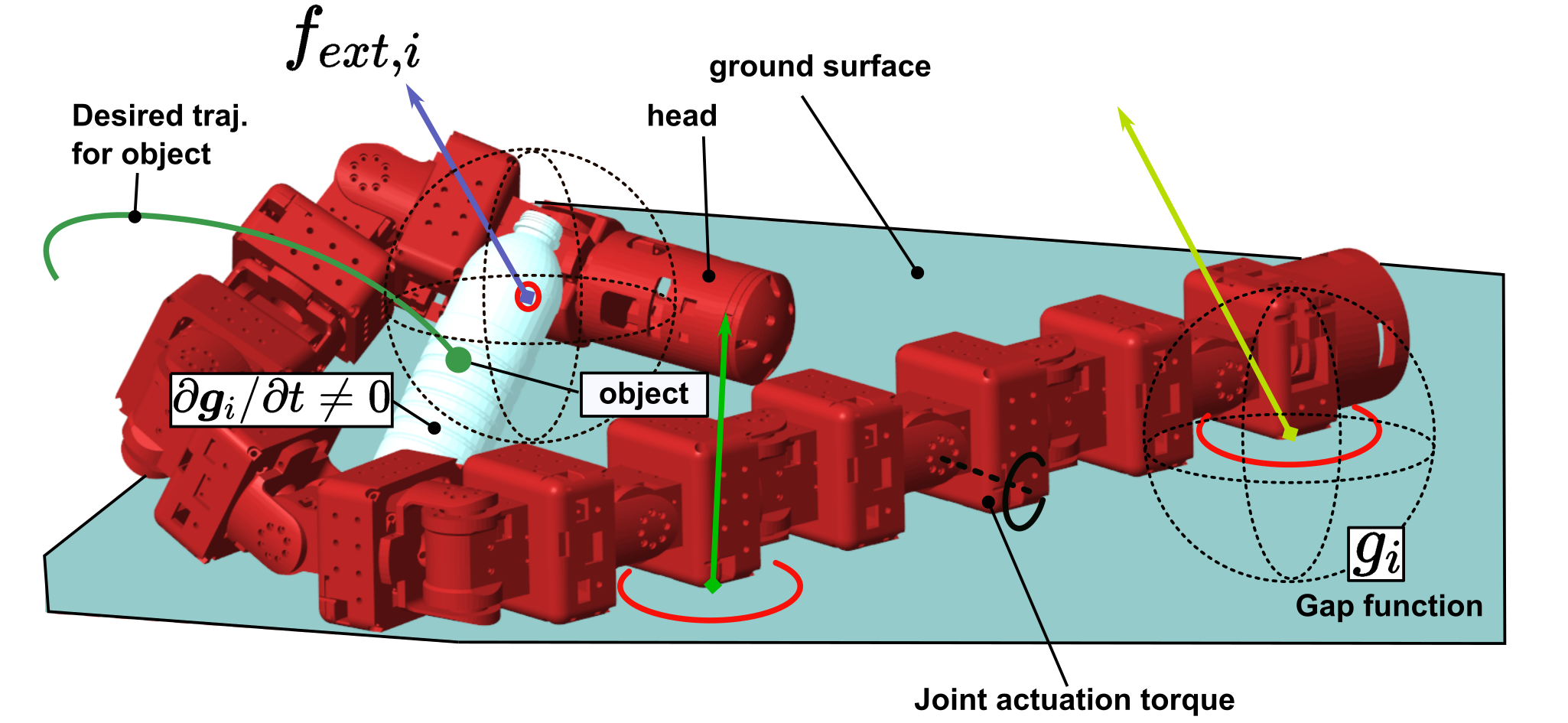}
    \caption{Full-dynamics model parameters in the object manipulation task considered in this paper}
    \label{fig:full-dyn}
\end{figure}

The dynamics governing the motion of the COBRA snake robot, possessing 11 body joints, are encapsulated in the following equations of motion:
\begin{equation}
    \begin{aligned}
        \bm M(\bm q) \dot{\bm u} - \bm h(\bm q, \bm u, \bm \tau) &= \sum_i \bm J_i^{\top}(\bm q) \bm f_{ext,i},\\
        \bm h(\bm q, \bm u, \bm \tau) &= \bm C(\bm q,\bm u)\bm u + \bm G(\bm q) + \bm B(\bm q)\bm \tau
    \end{aligned}
    \label{eq:eom}
\end{equation}
In this equation:
\begin{itemize}
    \item The mass-inertia matrix $\bm M$ pertains to a space of dimensionality $\mathbb{R}^{17 \times 17}$,
    \item The terms encompassing centrifugal, Coriolis, gravity, and actuation ($\bm \tau$) are succinctly represented by $\bm h \in \mathbb{R}^{17}$,
    \item External forces $\bm f_{ext,i}$ and their respective Jacobians $\bm J_i$ reside in the space $\mathbb{R}^{3 \times 17}$. 
\end{itemize}

\noindent For clarity and conciseness, details regarding the specific generalized coordinates and velocities of COBRA are omitted.

In the object manipulation problem under consideration and shown in Fig.~\ref{fig:full-dyn}, the external forces originate exclusively from active unilateral constraints, e.g., contact forces between the ground surface and the robot or between a movable object and the robot. This assumption conveniently establishes a complementarity relationship (i.e., slackness, where the product of two variables including force and displacement in the presence of holonomic constraints is zero) between the separation $\bm{g}_i$ (the gap between the body, terrain, and object) and the force exerted by a hard unilateral contact.

The concept of normal cone inclusion on the displacements, velocities, and acceleration levels allows for the expression \cite{studer_numerics_2009}:
\begin{equation}
    \begin{aligned}
        -\bm g_i & \in \partial \Psi_{i}\left(\bm f_{ext,i}\right) \equiv \mathcal{N}_{\mathcal{F}_i}\left(\bm f_{ext,i}\right) \\
        -\dot{\bm g}_i & \in \partial \Psi_{i}\left(\bm f_{ext,i}\right) \equiv \mathcal{N}_{\mathcal{F}_i}\left(\bm f_{ext,i}\right) \\
        -\ddot{\bm g}_i & \in \partial \Psi_{i}\left(\bm f_{ext,i}\right) \equiv \mathcal{N}_{\mathcal{F}_i}\left(\bm f_{ext,i}\right)
    \end{aligned}
    \label{eq:normal-cone-inclusion}
\end{equation}
where $\Psi_i(.)$ denotes the indicator function. The gap function \( \bm{g}_i \) is defined such that its total time derivative yields the relative constraint velocity \( \dot{\bm{g}}_i = \bm W_i^{\top} \bm u + \zeta_i \), where \( \bm W_i = \bm W_i(\bm q, t) = \left( \partial \bm{g}_i / \partial \bm q \right)^{\top} \) and \( \zeta_i = \zeta_i(\bm q, t) = \partial \bm{g}_i / \partial t \). 

Considering the primary objectives of loco-manipulation with COBRA, we examined various conditions of the normal cone inclusion as described in Eq.~\ref{eq:normal-cone-inclusion}. In scenarios where non-impulsive unilateral contact forces are employed to manipulate rigid objects (e.g., the box shown in Fig.~\ref{fig:full-dyn}) relative to the world, \( \partial \bm{g}_i / \partial t \neq 0 \). This factor holds significant importance in motion planning considered in this work and is enforced during optimization.

\begin{figure*}
    \centering
    \includegraphics[width=0.75\linewidth]{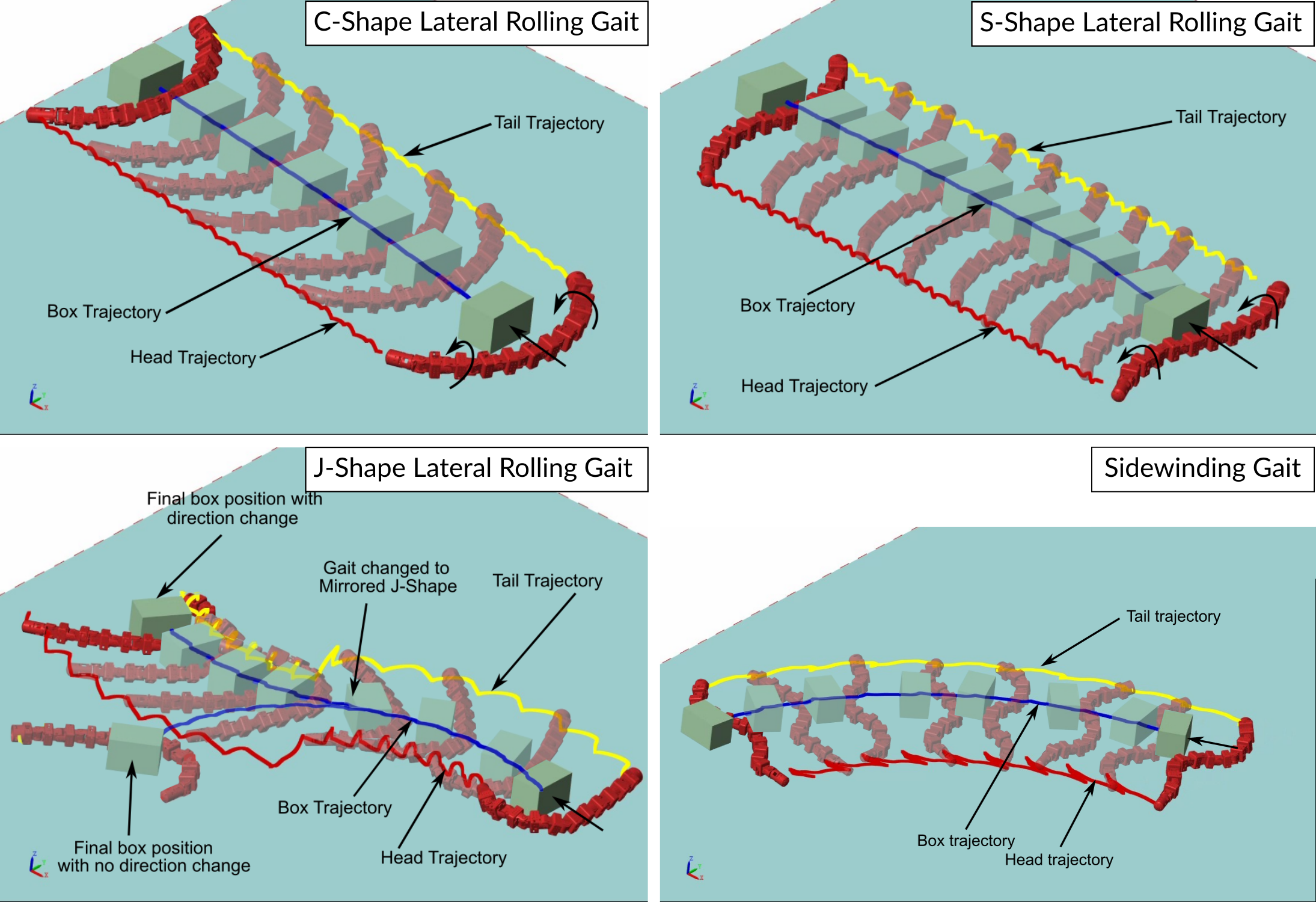}
    \caption{Snapshots depicting simulated forward box push utilizing various gaits executed in Matlab. The J-shape gait is an asymmetric variation of the C-gait, which allows changing the direction of movement of the box by mirroring the gait.}
    \label{fig:sim-snapshots}
\end{figure*}

The total time derivative of the relative constraint velocity yields the relative constraint accelerations $\ddot{\bm g}_i=\bm W^{\top} \dot{\bm u}+\hat{\zeta}_i$ where $\hat{\zeta}_i=\hat{\zeta}_i(\bm q, \bm u, t)$. We describe a geometric constraint on the acceleration level such that the initial conditions are fulfilled on velocity and displacement levels,
\begin{equation}
    \begin{aligned}
        &\bm g_i(\bm q, t)=0, \\
        &\dot{\bm g}_i=\bm W_i^\top \bm u+\zeta_i=0, \\
        &\ddot{\bm g}_i=\bm W_i^\top \dot{\bm u}+\hat \zeta_i=0,\\
        &\dot{\bm g}_i\left(\bm q_0, \bm u_0, t_0\right)=0,\\
        &\partial \bm{g}_i / \partial t \neq 0
    \end{aligned}   
\end{equation}
which means the generalized constraint forces must stand perpendicular to the manifolds $\bm g_i=0, \dot{\bm g_i}=0$ and $\ddot{\bm g}_i=0$. This formulation directly accommodates the integration of friction laws, which naturally pertain to velocity considerations. We dissect the contact forces into normal and tangential components, denoted as $\bm f_{ext,i}=\left[\begin{array}{ll}f_{N,i}, & \bm f_{T,i}^{\top}\end{array}\right]^{\top} \in \mathcal{F}_{i}$. 

In this context, the force space $\mathcal{F}_i$ facilitate the specification of non-negative normal forces ($\mathbb{R}_{0}^+$) and tangential forces adhering to Coulomb friction $\left\{\bm f_{T,i} \in \mathbb{R}^2, ||\bm f_{T,i}||<\mu\left|f_{N,i}\right| \right\}$, with $\mu$ representing the friction coefficient.

The underlying rationale behind this approach is that while the force remains confined within the interior of its designated subspace, the contact velocity remains constrained to zero. Conversely, non-zero gap velocities only arise when the forces reach the boundary of their permissible set, indicating either a zero normal force or the maximum friction force opposing the direction of motion.

To proceed in the loco-manipulation problem considered here, it proves advantageous to reconfigure Eq.~\ref{eq:eom} into local contact coordinates (task space). This can be accomplished by recognizing the relationship:
\begin{equation}
    \dot{\bm g}=\bm J_{\mathrm{c}} \bm u,     
\end{equation}
where $\bm g$ and $\bm J_c$ represent the stacked contact separations and Jacobians, respectively. By differentiating the above equation with respect to time and substituting Eq.~\ref{eq:eom}, we obtain
\begin{equation}
    \ddot{\bm g} = \bm J_c \bm M^{-1} \bm J_c^{\top} \bm f_{ext} + \dot{\bm J}_c \bm u + \bm J_c \bm M^{-1} \bm h,
\end{equation} 
where $\bm G = \bm J_c \bm M^{-1} \bm J_c^{\top}$ -- the Delassus matrix -- signifies the apparent inverse inertia at the contact points, and $ \bm c = \dot{\bm J}_c \bm u + \bm J_c \bm M^{-1} \bm h$ encapsulates all terms independent of the stacked external forces $\bm f_{ext}$. At this juncture, the principle of least action asserts that the contact forces are determined by the solution of the constrained optimization problem:
\begin{equation}
    \left\{\begin{aligned}
        &\underset{\left\{\bm f_{ext,i},\bm u\right\}}{\operatorname{minimize}} \frac{1}{2} \bm f_{ext}^\top \bm G \bm f_{ext}+\bm f_{ext}^\top \bm c \\
        &{\operatorname{s.t.}}\\
        &{\operatorname{(1)} \bm M(\bm q) \dot{\bm u} - \bm h(\bm q, \bm u, \bm \tau) - \sum_i \bm J_i^{\top}(\bm q) \bm f_{ext,i}}=0\\
        &{\operatorname{(2)}} \|\bm q\|\leq q_{max}\\
        &{\operatorname{(3)}} \|\bm \tau\|\leq \tau_{max}\\
    \end{aligned}\right.
\end{equation}
where $q_{max}$, and $\tau_{max}$ denote maximum joint movements, and actuation toruqes, respectively. In the above optimization problem, (1), (2), and (3) enforce dynamics agreement, kinematics restrictions, and actution saturations, respectively. Next, a time-stepping methodology facilitates the integration of system dynamics across a time interval $\Delta t$ while internally addressing the resolution of contact forces. We take the shooting method to find optimal $\bm u_{ref}$ for minimized joint torques $\bm \tau$ such that generalized contact forces $\bm f_{ext}$ stand orthogonal to gap functions and their derivative.

\begin{figure*}
    \centering    \includegraphics[width=0.8\linewidth]{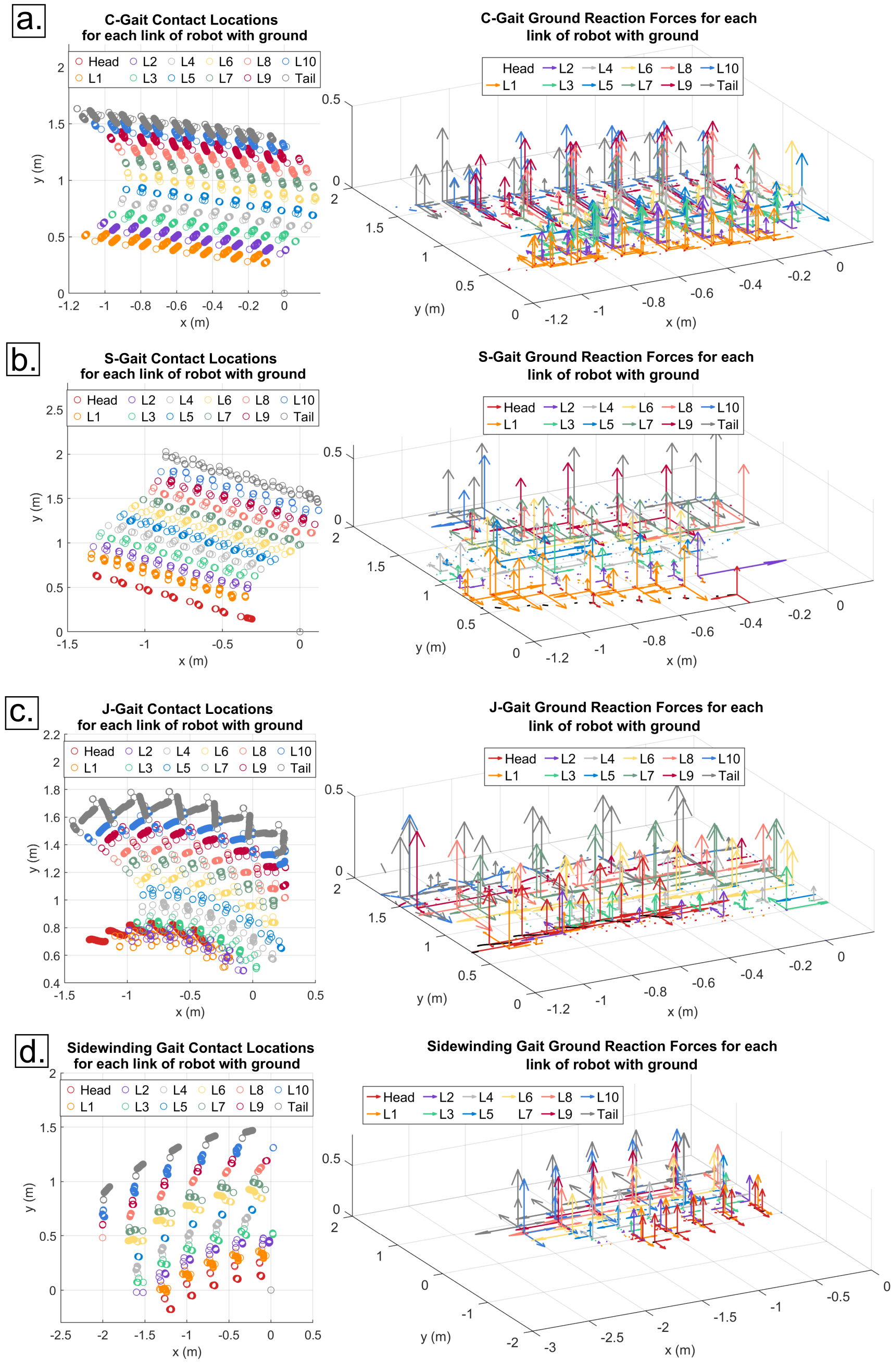}
    \caption{This image depicts the contact points and unilateral ground reaction forces during (a): C-shaped gaits, (b) S-Shaped Gait, (C) J-Shaped Gait, (d) Sidewinding gait, performed by the high-fidelity COBRA model simulated in the MATLAB environment. The contact forces consist of tangential forces ($\bm f_{T,i}$) and normal forces ($f_{N,i}$). L$x$ here refers to Link number $x$ on the robot, numbered from 1-10 starting from the head.}
    \label{fig:Contact-points-robot}
\end{figure*}

\begin{figure*}
    \centering    \includegraphics[width=0.9\linewidth]{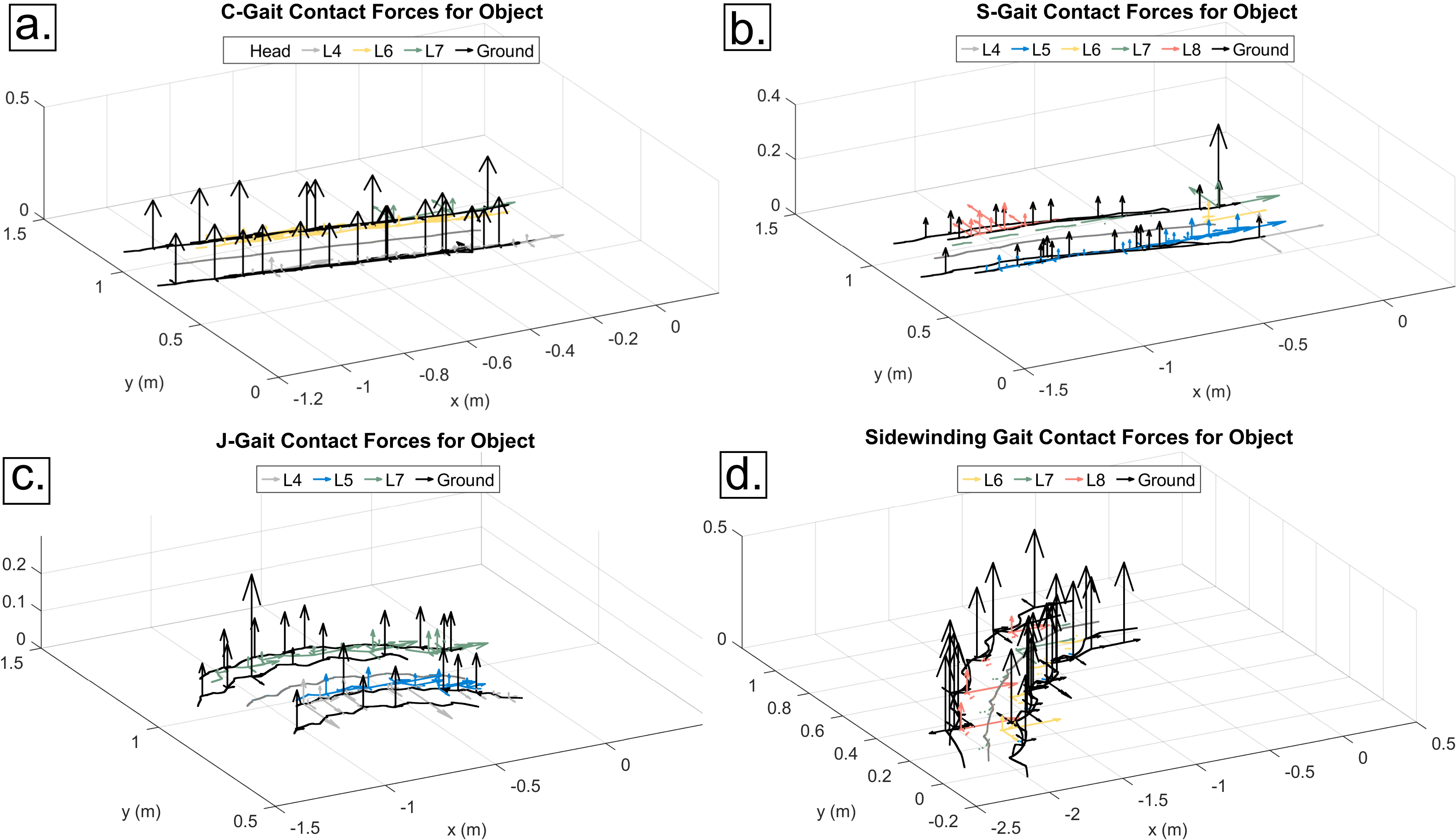}
    \caption{This image depicts the contact points and unilateral ground reaction forces for contact interactions of the manipulated object with the robot and with the ground during (a): C-shaped gaits, (b) S-Shaped Gait, (C) J-Shaped Gait, (d) Sidewinding gait, performed by the high-fidelity COBRA model simulated in the MATLAB environment. The contact forces consist of tangential forces ($\bm f_{T,i}$) and normal forces ($f_{N,i}$). L$x$ here refers to Link number $x$ on the robot, numbered from 1-10 starting from the head.}
    \label{fig:Contact-points-box}
\end{figure*}

\begin{figure}
    \centering    \includegraphics[width=0.7\linewidth]{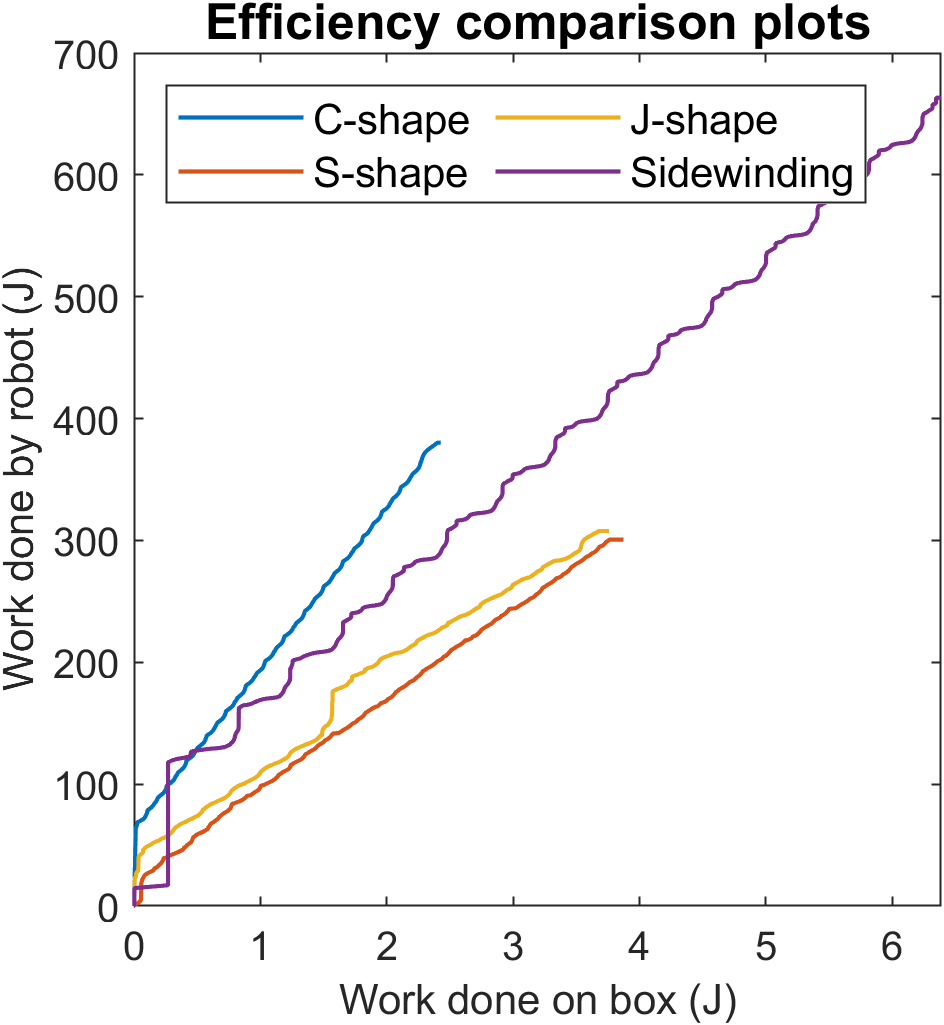}
    \caption{Depicts the efficiency of each gait based on the total work done by the robot for locomotion against total work done on the box. Larger slope indicates a less efficient gait as more work is done on locomotion in return for smaller work done on the box.}
    \label{fig:Work efficiency plots}
\end{figure}

\begin{figure}
    \centering    
    \includegraphics[width=0.7\linewidth]{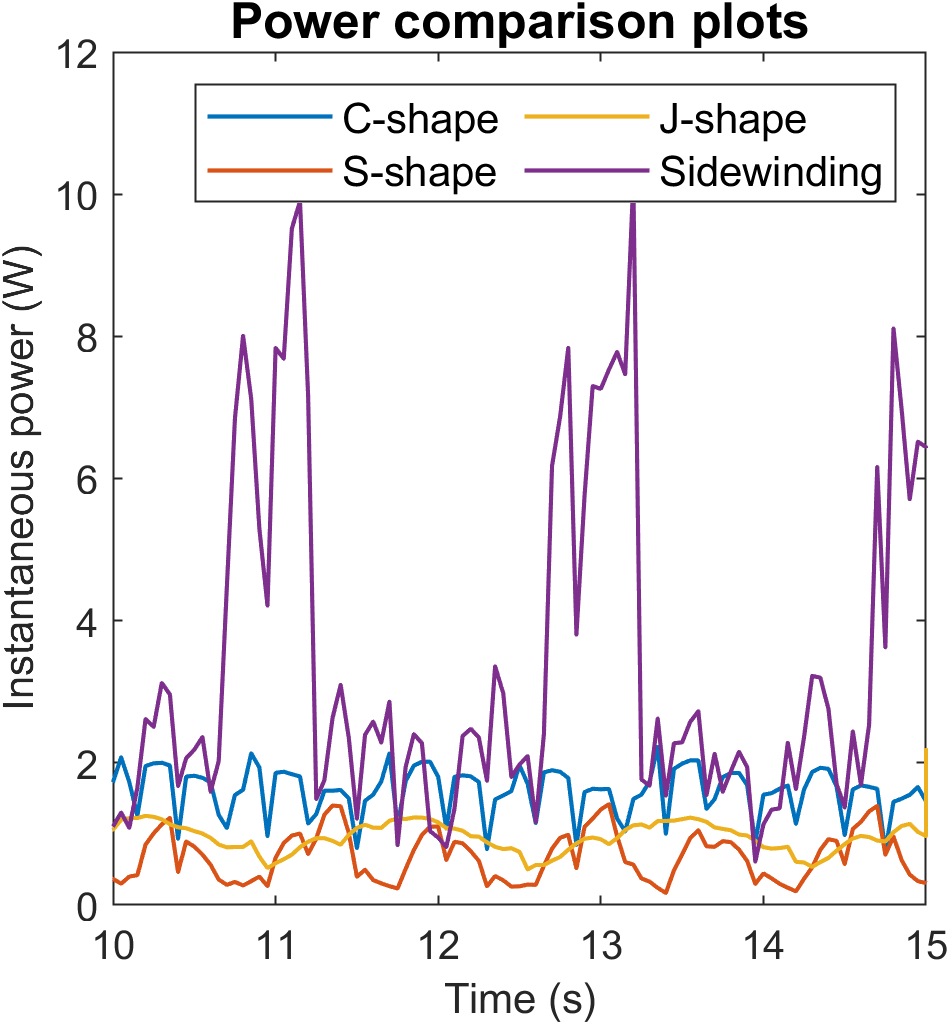}
    \caption{Shows the instantaneous power consumption by the robot to execute locomotion for each of the considered gaits.}
    \label{fig:Instantaneous power plots}
\end{figure}

\begin{figure}
    \centering    \includegraphics[width=0.9\linewidth]{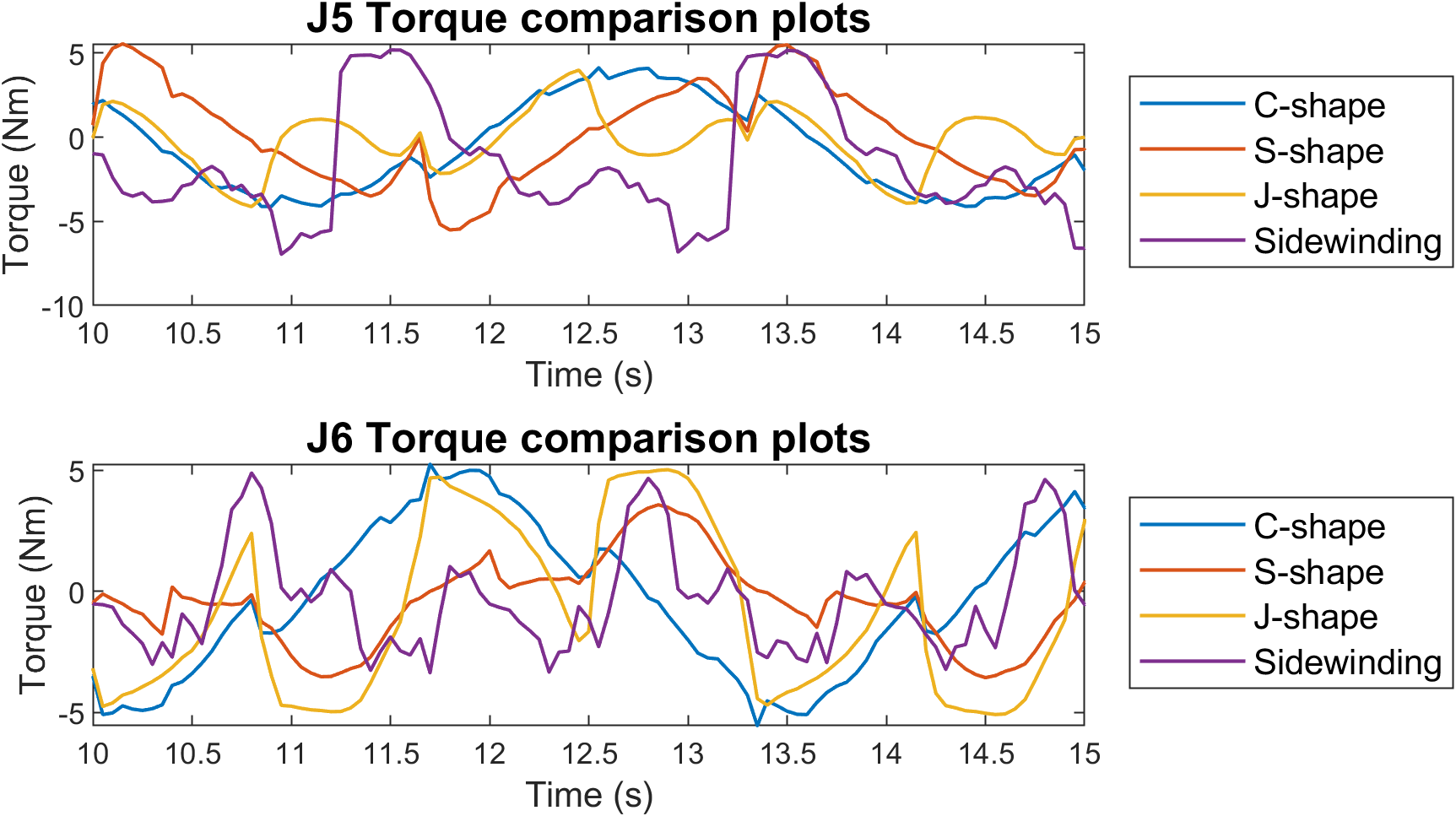}
    \caption{Depicts the torque profile for the central yawing (J6) and pitching (J5) joints for each gait. Other joints show similar profiles offset by a phase angle based on the executed gait.}
    \label{fig:Torque plots}
\end{figure}

\begin{figure}
    \centering    \includegraphics[width=0.7\linewidth]{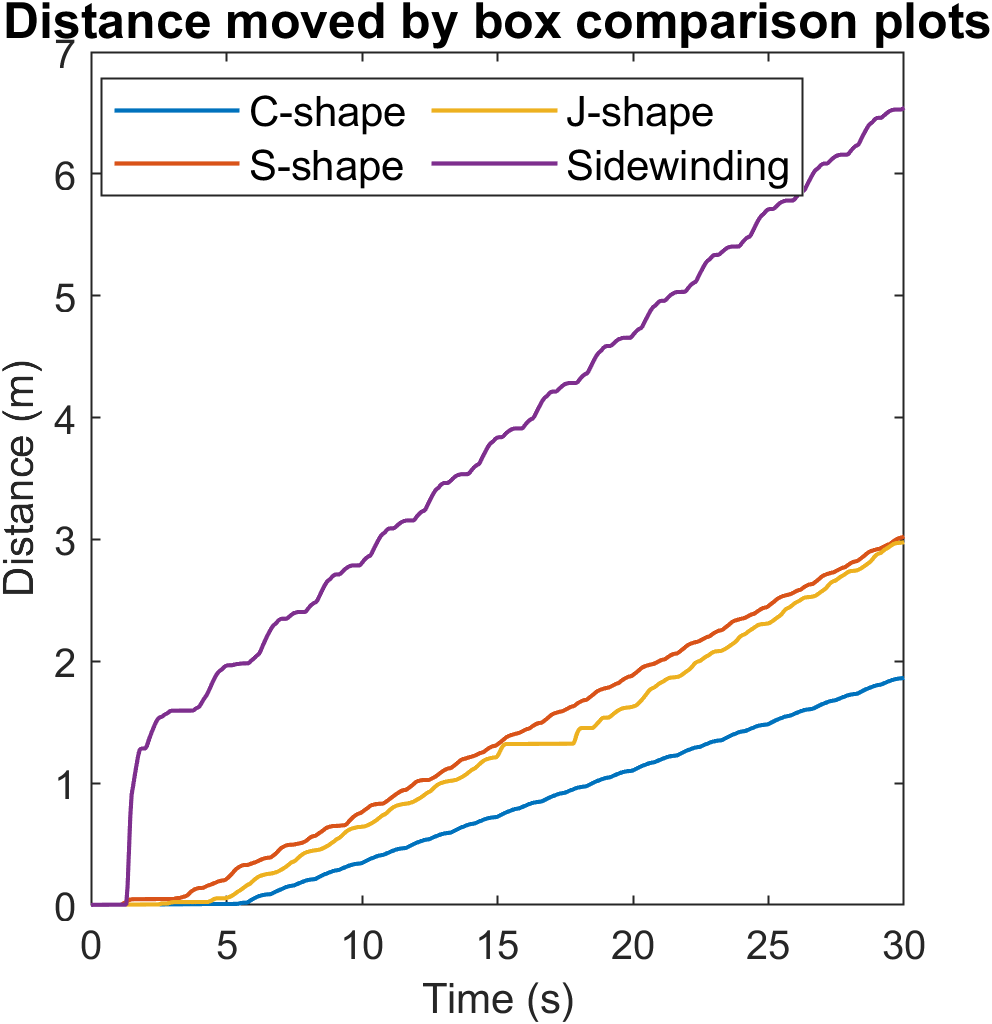}
    \caption{Depicts the total distance the robot moves the box for each gait in the same amount of time.}
    \label{fig:Distance comparison plots}
\end{figure}

\section{Simulation Setup}
\label{sec:sim-setup}
A high fidelity simulation has been created using the MATLAB \textit{Simulink Multibody Toolbox}. Each link is modeled as a rigid body weighing $0.5$ kg with inertia matrices derived automatically by MATLAB from the geometry assuming homogeneous mass distribution. The inertia tensor for each of the ten identical body links is ($\mathbf{I}_{xx} = 7.167\e{-4},~\mathbf{I}_{yy} = 8.704\e{-4},~\mathbf{I}_{zz} = 8.626\e{-4}~ \text{kgm}^2$), and the inertia tensors for the head and tail modules are ($\mathbf{I}_{xx} - 4.4562\e{-4},~\mathbf{I}_{yy} = 1.710\e{-3},~\mathbf{I}_{zz} = 1.793\e{-3}~ \text{kgm}^2$) and ($\mathbf{I}_{xx} = 8.182\e{-4},~\mathbf{I}_{yy} = 1.141\e{-3},~\mathbf{I}_{zz} = 1.109\e{-3}~ \text{kgm}^2$) along the primary axes. The links are connected through a position controlled revolute joint with axis and range of motion mimicking the real robot. The object being manipulated is modeled as a solid box of weight $0.5$ kg. The normal forces for all contact interactions between robot links, ground surface and object are modeled using a smooth spring-damper model with spring stiffness $1\e{-4}~N/m$ and damping coefficient $1\times10^3~Ns/m$. Friction forces are modeled using a smooth stick-slip model with coefficient of static friction of $1.3$, coefficient of dynamic friction of $1.0$ and critical velocity $1\e{-3}~m/s$. The dynamics are solved using MATLAB's \textit{ode45} with a fixed timestep of $1\e{-4}$ seconds. 

\section{Results}
\label{sec:results}

We performed numerical dynamics integration of COBRA and a box interactions on flat ground. In Fig.~\ref{fig:sim-snapshots}, snapshots illustrating simulated forward box push using C-, S-, J-shaped lateral rolling, and sidewinding gait are presented. In these simulations, the goal is to move the box on the flat ground towards a specified point, and corresponding suitable joint commands are derived. The J-shaped lateral rolling gait is asymmetrical and can be mirrored to execute control on the direction in which the object is moved. The composite Fig.~\ref{fig:Contact-points-robot} illustrates the contact points and unilateral ground reaction forces during (a) C-shaped gait, (b) S-shaped gait, (c) J-shaped gait, and (d) sidewinding gait, executed by the high-fidelity COBRA model simulated in the MATLAB environment. The contact forces encompass tangential forces ($\bm f_{T,i}$) and normal forces ($f_{N,i}$), where $Lx$ denotes Link number $x$ on the robot, numbered from 1 to 10 starting from the head. 

In the composite Fig.~\ref{fig:Contact-points-box}, we depict the contact points and unilateral ground reaction forces for contact interactions of the manipulated object with the robot and with the ground during (a) C-shaped gait, (b) S-shaped gait, (c) J-shaped gait, and (d) sidewinding gait. Figures \ref{fig:Work efficiency plots} and \ref{fig:Instantaneous power plots} compare the efficiency of the four gaits in performing the prescribed task of moving the box. Figure \ref{fig:Work efficiency plots} plots the work done internally by the robot for locomotion against the work done by the robot on the box. The joint torques $\bm \tau$ for the gaits are shown in Fig. \ref{fig:Torque plots}. 

More efficient gaits would need less work in locomotion to do more work on the box. In this respect, the S and J shape lateral rolling gaits perform more efficiently than other gaits, with Sidewinding doing the most work on the box, but consuming the most energy to do it. This is corroborated by the plot of instantaneous power for the robot for locomotion in Fig.~\ref{fig:Instantaneous power plots} that show the large peaks in instantaneous power from sidewinding, as compared to the more steady power consumption by other gaits. This results in a slower but more energy efficient loco-manipulation. Figure \ref{fig:Distance comparison plots} shows the relative distance moved by the box as a result of each gait operating for the same amount of time.






\section{Concluding Remarks}
\label{sec:conclusion}

This paper introduces an optimization-based approach to path planning and object manipulation for loco-manipulation, employing a morpho-functional robot named COBRA, equipped with manifold moving joints and contact-rich behavior. COBRA features 11 actuated joints and onboard sensing and computation capabilities. We develop a high-fidelity model of COBRA to manipulate a box on a flat surface, utilizing non-impulsive implicit contact path planning. Our simulation results demonstrate successful object manipulation, accompanied by an analysis of ground reaction forces and intermittent contact locations.

In future research, we intend to leverage motion capture and tactile sensors (currently undergoing integration) to conduct real-world experiments. With the integration of contact force estimation into the hardware, our contact-implicit optimization approach will be applied to the physical robot. This will allow us to showcase loco-manipulation in real-world scenarios, such as pushing objects on ramps or anchoring the robot's body to the ground to lift and place objects on elevated platforms.

\nocite{salagame_how_2023, jiang_hierarchical_2023, salagame_letter_2022, sihite_optimization-free_2021-1, sihite_multi-modal_2023, salagame_quadrupedal_2023, sihite_optimization-free_2021, sihite_unilateral_2021-1, liang_rough-terrain_2021-1}

\printbibliography

@book{studer_numerics_2009,
	address = {Berlin, Heidelberg},
	series = {Lecture {Notes} in {Applied} and {Computational} {Mechanics}},
	title = {Numerics of {Unilateral} {Contacts} and {Friction}},
	volume = {47},
	isbn = {978-3-642-01099-6 978-3-642-01100-9},
	url = {http://link.springer.com/10.1007/978-3-642-01100-9},
	urldate = {2024-02-18},
	publisher = {Springer},
	author = {Studer, Christian},
	editor = {Pfeiffer, Friedrich and Wriggers, Peter},
	year = {2009},
	doi = {10.1007/978-3-642-01100-9},
}

@article{burdick_sidewinding_1994,
	title = {A 'sidewinding' locomotion gait for hyper-redundant robots},
	volume = {9},
	issn = {0169-1864},
	url = {https://doi.org/10.1163/156855395X00166},
	doi = {10.1163/156855395X00166},
	number = {3},
	urldate = {2023-01-11},
	journal = {Advanced Robotics},
	author = {Burdick, J.W. and Radford, J. and Chirikjian, G.S.},
	month = jan,
	year = {1994},
	note = {Publisher: Taylor \& Francis
\_eprint: https://doi.org/10.1163/156855395X00166},
	pages = {195--216},
}

@inproceedings{yim_new_1994,
	title = {New locomotion gaits},
	doi = {10.1109/ROBOT.1994.351134},
	booktitle = {Proceedings of the 1994 {IEEE} {International} {Conference} on {Robotics} and {Automation}},
	author = {Yim, M.},
	month = may,
	year = {1994},
	pages = {2508--2514 vol.3},
}

@article{liljeback_modular_2005,
	series = {16th {IFAC} {World} {Congress}},
	title = {{MODULAR} {PNEUMATIC} {SNAKE} {ROBOT} {3D} {MODELLING}, {IMPLEMENTATION} {AND} {CONTROL}},
	volume = {38},
	issn = {1474-6670},
	url = {https://www.sciencedirect.com/science/article/pii/S147466701637286X},
	doi = {10.3182/20050703-6-CZ-1902.01274},
	language = {en},
	number = {1},
	urldate = {2023-01-11},
	journal = {IFAC Proceedings Volumes},
	author = {Liljebäck, Pål and Stavdahl, Øyvind and Pettersen, Kristin Y.},
	month = jan,
	year = {2005},
	pages = {19--24},
}

@inproceedings{ohno_design_2001,
	title = {Design of slim slime robot and its gait of locomotion},
	volume = {2},
	url = {https://ieeexplore.ieee.org/abstract/document/976252},
	doi = {10.1109/IROS.2001.976252},
	urldate = {2024-02-18},
	booktitle = {Proceedings 2001 {IEEE}/{RSJ} {International} {Conference} on {Intelligent} {Robots} and {Systems}. {Expanding} the {Societal} {Role} of {Robotics} in the the {Next} {Millennium} ({Cat}. {No}.{01CH37180})},
	author = {Ohno, H. and Hirose, S.},
	month = oct,
	year = {2001},
	pages = {707--715 vol.2},
}

@article{rincon_ver-vite_2003,
	title = {Ver-vite: dynamic and experimental analysis for inchwormlike biomimetic robots},
	volume = {10},
	issn = {1558-223X},
	shorttitle = {Ver-vite},
	url = {https://ieeexplore.ieee.org/abstract/document/1256298?casa_token=k__nqVjXa2gAAAAA:fbCJ59MTBG7V9S5SsdSBFF4Gh_edPWk9eLySqUIwI0o2Zo6U4UbSviX-YyjBLLoRf41KFCdx4sw},
	doi = {10.1109/MRA.2003.1256298},
	number = {4},
	urldate = {2024-02-18},
	journal = {IEEE Robotics \& Automation Magazine},
	author = {Rincon, D.M. and Sotelo, J.},
	month = dec,
	year = {2003},
	note = {Conference Name: IEEE Robotics \& Automation Magazine},
	pages = {53--57},
}

@inproceedings{ma_analysis_1999,
	title = {Analysis of snake movement forms for realization of snake-like robots},
	volume = {4},
	url = {https://ieeexplore.ieee.org/abstract/document/774054?casa_token=Gsc2cQ4ucrYAAAAA:WjWzMJaMyvuGGnBo5eHec1VjnzcicU417fqE_I3Y4pyGrCwdIgHodQm088rPhTvADaMEf6SzmIU},
	doi = {10.1109/ROBOT.1999.774054},
	urldate = {2024-02-18},
	booktitle = {Proceedings 1999 {IEEE} {International} {Conference} on {Robotics} and {Automation} ({Cat}. {No}.{99CH36288C})},
	author = {Ma, S.},
	month = may,
	year = {1999},
	note = {ISSN: 1050-4729},
	pages = {3007--3013 vol.4},
}

@inproceedings{ma_analysis_2003,
	title = {Analysis of creeping locomotion of a snake robot on a slope},
	volume = {2},
	url = {https://ieeexplore.ieee.org/abstract/document/1241899?casa_token=dSZSrNOb-fUAAAAA:CLEJQoSQaNVKiL0W45PrBamuEDlIxqiG2mYDSZ9IMdh4qifsU0SyywLR7fqHngxGVF6MYIy7L-c},
	doi = {10.1109/ROBOT.2003.1241899},
	urldate = {2024-02-18},
	booktitle = {2003 {IEEE} {International} {Conference} on {Robotics} and {Automation} ({Cat}. {No}.{03CH37422})},
	author = {Ma, Shugen and Tadokoro, N. and Li, Bin and Inoue, K.},
	month = sep,
	year = {2003},
	note = {ISSN: 1050-4729},
	pages = {2073--2078 vol.2},
}

@inproceedings{wiriyacharoensunthorn_analysis_2002,
	title = {Analysis and design of a multi-link mobile robot ({Serpentine})},
	volume = {2},
	url = {https://ieeexplore.ieee.org/abstract/document/1189249?casa_token=BxdHGxRUgHAAAAAA:Hy_5TRm4yTg7k8ade4N6w4TOswIGftGocpUgB_SqLf4b0EwZkpD9C_U1vedIsGoeczWbDXzOM6Q},
	doi = {10.1109/ICIT.2002.1189249},
	urldate = {2024-02-18},
	booktitle = {2002 {IEEE} {International} {Conference} on {Industrial} {Technology}, 2002. {IEEE} {ICIT} '02.},
	author = {Wiriyacharoensunthorn, P. and Laowattana, S.},
	month = dec,
	year = {2002},
	pages = {694--699 vol.2},
}

@misc{salagame_quadrupedal_2023,
	title = {Quadrupedal {Locomotion} {Control} {On} {Inclined} {Surfaces} {Using} {Collocation} {Method}},
	url = {http://arxiv.org/abs/2312.08621},
	doi = {10.48550/arXiv.2312.08621},
	urldate = {2024-01-17},
	publisher = {arXiv},
	author = {Salagame, Adarsh and Gianello, Maria and Wang, Chenghao and Venkatesh, Kaushik and Pitroda, Shreyansh and Rajput, Rohit and Sihite, Eric and Leeser, Miriam and Ramezani, Alireza},
	month = dec,
	year = {2023},
	note = {arXiv:2312.08621 [cs, eess]},
}

@misc{jiang_hierarchical_2023,
	title = {Hierarchical {RL}-{Guided} {Large}-scale {Navigation} of a {Snake} {Robot}},
	url = {http://arxiv.org/abs/2312.03223},
	doi = {10.48550/arXiv.2312.03223},
	urldate = {2024-01-17},
	publisher = {arXiv},
	author = {Jiang, Shuo and Salagame, Adarsh and Ramezani, Alireza and Wong, Lawson},
	month = dec,
	year = {2023},
	note = {arXiv:2312.03223 [cs, eess]},
}

@misc{salagame_how_2023,
	title = {How {Strong} a {Kick} {Should} be to {Topple} {Northeastern}'s {Tumbling} {Robot}?},
	url = {http://arxiv.org/abs/2311.14878},
	doi = {10.48550/arXiv.2311.14878},
	urldate = {2023-12-10},
	publisher = {arXiv},
	author = {Salagame, Adarsh and Bhattachan, Neha and Caetano, Andre and McCarthy, Ian and Noyes, Henry and Petersen, Brandon and Qiu, Alexander and Schroeter, Matthew and Smithwick, Nolan and Sroka, Konrad and Widjaja, Jason and Bohra, Yash and Venkatesh, Kaushik and Gangaraju, Kruthika and Ghanem, Paul and Mandralis, Ioannis and Sihite, Eric and Kalantari, Arash and Ramezani, Alireza},
	month = nov,
	year = {2023},
	note = {arXiv:2311.14878 [cs, eess]},
}

@inproceedings{sihite_efficient_2022,
	title = {Efficient {Path} {Planning} and {Tracking} for {Multi}-{Modal} {Legged}-{Aerial} {Locomotion} {Using} {Integrated} {Probabilistic} {Road} {Maps} ({PRM}) and {Reference} {Governors} ({RG})},
	url = {https://ieeexplore.ieee.org/abstract/document/9992754?casa_token=E0v5fTczBNsAAAAA:23fuUCarpuZwI_Fn00Mg5JZjwxGq8BB-i5n2y-rXXLdp14E3esNoXewTLtycARuiLaW-lmxIbWs},
	doi = {10.1109/CDC51059.2022.9992754},
	urldate = {2023-12-09},
	booktitle = {2022 {IEEE} 61st {Conference} on {Decision} and {Control} ({CDC})},
	author = {Sihite, Eric and Mottis, Benjamin and Ghanem, Paul and Ramezani, Alireza and Gharib, Morteza},
	month = dec,
	year = {2022},
	note = {ISSN: 2576-2370},
	pages = {764--770},
}

@article{sihite_multi-modal_2023,
	title = {Multi-{Modal} {Mobility} {Morphobot} ({M4}) with appendage repurposing for locomotion plasticity enhancement},
	volume = {14},
	copyright = {2023 The Author(s)},
	issn = {2041-1723},
	url = {https://www.nature.com/articles/s41467-023-39018-y},
	doi = {10.1038/s41467-023-39018-y},
	language = {en},
	number = {1},
	urldate = {2023-11-22},
	journal = {Nature Communications},
	author = {Sihite, Eric and Kalantari, Arash and Nemovi, Reza and Ramezani, Alireza and Gharib, Morteza},
	month = jun,
	year = {2023},
	note = {Number: 1
Publisher: Nature Publishing Group},
	pages = {3323},
}

@article{dangol_control_2021,
	title = {Control of {Thruster}-{Assisted}, {Bipedal} {Legged} {Locomotion} of the {Harpy} {Robot}},
	volume = {8},
	issn = {2296-9144},
	url = {https://www.frontiersin.org/articles/10.3389/frobt.2021.770514},
	urldate = {2023-11-18},
	journal = {Frontiers in Robotics and AI},
	author = {Dangol, Pravin and Sihite, Eric and Ramezani, Alireza},
	year = {2021},
}

@inproceedings{sihite_optimization-free_2021,
	title = {Optimization-free {Ground} {Contact} {Force} {Constraint} {Satisfaction} in {Quadrupedal} {Locomotion}},
	doi = {10.1109/CDC45484.2021.9683155},
	booktitle = {2021 60th {IEEE} {Conference} on {Decision} and {Control} ({CDC})},
	author = {Sihite, Eric and Dangol, Pravin and Ramezani, Alireza},
	month = dec,
	year = {2021},
	note = {ISSN: 2576-2370},
	pages = {713--719},
}

@inproceedings{liang_rough-terrain_2021-1,
	title = {Rough-{Terrain} {Locomotion} and {Unilateral} {Contact} {Force} {Regulations} {With} a {Multi}-{Modal} {Legged} {Robot}},
	doi = {10.23919/ACC50511.2021.9483189},
	booktitle = {2021 {American} {Control} {Conference} ({ACC})},
	author = {Liang, Kaier and Sihite, Eric and Dangol, Pravin and Lessieur, Andrew and Ramezani, Alireza},
	month = may,
	year = {2021},
	note = {ISSN: 2378-5861},
	pages = {1762--1769},
}

@inproceedings{sihite_optimization-free_2021-1,
	title = {Optimization-free {Ground} {Contact} {Force} {Constraint} {Satisfaction} in {Quadrupedal} {Locomotion}},
	doi = {10.1109/CDC45484.2021.9683155},
	booktitle = {2021 60th {IEEE} {Conference} on {Decision} and {Control} ({CDC})},
	author = {Sihite, Eric and Dangol, Pravin and Ramezani, Alireza},
	month = dec,
	year = {2021},
	note = {ISSN: 2576-2370},
	pages = {713--719},
}

@inproceedings{sihite_unilateral_2021-1,
	title = {Unilateral {Ground} {Contact} {Force} {Regulations} in {Thruster}-{Assisted} {Legged} {Locomotion}},
	doi = {10.1109/AIM46487.2021.9517648},
	booktitle = {2021 {IEEE}/{ASME} {International} {Conference} on {Advanced} {Intelligent} {Mechatronics} ({AIM})},
	author = {Sihite, Eric and Dangol, Pravin and Ramezani, Alireza},
	month = jul,
	year = {2021},
	note = {ISSN: 2159-6255},
	pages = {389--395},
}

@misc{salagame_letter_2022,
	title = {A {Letter} on {Progress} {Made} on {Husky} {Carbon}: {A} {Legged}-{Aerial}, {Multi}-modal {Platform}},
	shorttitle = {A {Letter} on {Progress} {Made} on {Husky} {Carbon}},
	url = {http://arxiv.org/abs/2207.12254},
	doi = {10.48550/arXiv.2207.12254},
	urldate = {2023-05-17},
	publisher = {arXiv},
	author = {Salagame, Adarsh and Manjikian, Shoghair and Wang, Chenghao and Krishnamurthy, Kaushik Venkatesh and Pitroda, Shreyansh and Gupta, Bibek and Jacob, Tobias and Mottis, Benjamin and Sihite, Eric and Ramezani, Milad and Ramezani, Alireza},
	month = jul,
	year = {2022},
	note = {arXiv:2207.12254 [cs, eess]},
}

@article{shan_design_1993,
	title = {Design and motion planning of a mechanical snake},
	volume = {23},
	issn = {2168-2909},
	doi = {10.1109/21.247890},
	number = {4},
	journal = {IEEE Transactions on Systems, Man, and Cybernetics},
	author = {Shan, Y. and Koren, Y.},
	month = jul,
	year = {1993},
	note = {Conference Name: IEEE Transactions on Systems, Man, and Cybernetics},
	pages = {1091--1100},
}

@incollection{ramezani_atrias_2012,
	title = {Atrias 2.0, a new 3d bipedal robotic walker and runner},
	isbn = {978-981-4415-94-1},
	url = {https://www.worldscientific.com/doi/abs/10.1142/9789814415958_0060},
	urldate = {2023-05-17},
	booktitle = {Adaptive {Mobile} {Robotics}},
	publisher = {WORLD SCIENTIFIC},
	author = {Ramezani, Alireza and Grizzle, J.w.},
	month = may,
	year = {2012},
	doi = {10.1142/9789814415958_0060},
	pages = {467--474},
}

\end{document}